\documentclass{article}

\usepackage{microtype}
\usepackage{graphicx}
\usepackage{subfigure}
\usepackage{booktabs} 
\usepackage{tikz}
\usepackage{wrapfig}

\usepackage{booktabs}    
\usepackage{multirow}    
\usepackage{array}       
\usepackage{siunitx}     

\usepackage{hyperref}

\usepackage[accepted]{icml2025}

\usepackage{amsmath}
\usepackage{amssymb}
\usepackage{mathtools}
\usepackage{amsthm}

\newtheorem{assum}{Assumption}

\newcommand{\indep}{\perp \!\!\! \perp}

\DeclareMathOperator*{\argmin}{arg\,min}

\usetikzlibrary{shapes.geometric, shapes,decorations,arrows,calc,arrows.meta,fit,positioning,
backgrounds}
\tikzset{
    -Latex,auto,node distance =1 cm and 1 cm,semithick,
    state/.style ={ellipse, draw, minimum width = 0.7 cm},
    point/.style = {circle, draw, inner sep=0.04cm,fill,node contents={}},
    bidirected/.style={Latex-Latex,dashed},
    el/.style = {inner sep=2pt, align=left, sloped}
}

\usepackage[capitalize,noabbrev]{cleveref}

\theoremstyle{plain}

\theoremstyle{definition}

\theoremstyle{remark}

\usepackage[textsize=tiny]{todonotes}

\icmltitlerunning{DAG-aware Transformer for Causal Effect Estimation}

\begin{document}

\twocolumn[
\icmltitle{DAG-aware Transformer for Causal Effect Estimation }



\icmlsetsymbol{equal}{*}

\begin{icmlauthorlist}
\icmlauthor{Manqing Liu}{yyy,sch}
\icmlauthor{David R. Bellamy}{comp}
\icmlauthor{Andrew L. Beam}{yyy,comp}
\end{icmlauthorlist}

\icmlaffiliation{yyy}{Department of Epidemiology, CAUSALab, Harvard University, Boston, USA}
\icmlaffiliation{sch}{School of Engineering and Applied Sciences, Harvard University, Cambridge, USA}
\icmlaffiliation{comp}{Lila Sciences, Cambridge, USA}

\icmlcorrespondingauthor{Andrew Beam}{andrew\_beam@hms.harvard.edu}
\icmlkeywords{Machine Learning, causality, DAG, transformer, ICML}

\vskip 0.3in
]



\printAffiliationsAndNotice{}  

\begin{abstract}
Causal inference is a critical task across fields such as healthcare, economics, and the social sciences. While recent advances in machine learning, especially those based on the deep-learning architectures, have shown potential in estimating causal effects, existing approaches often fall short in handling complex causal structures and lack adaptability across various causal scenarios. In this paper, we present a novel transformer-based method for causal inference that overcomes these challenges. The core innovation of our model lies in its integration of causal Directed Acyclic Graphs (DAGs) directly into the attention mechanism, enabling it to accurately model the underlying causal structure. This allows for flexible estimation of both average treatment effects (ATE) and conditional average treatment effects (CATE). Extensive experiments on both synthetic and real-world datasets demonstrate that our approach surpasses existing methods in estimating causal effects across a wide range of scenarios. The flexibility and robustness of our model make it a valuable tool for researchers and practitioners tackling complex causal inference problems. Our code is available at \url{https://github.com/ManqingLiu/DAGawareTransformer}.
\end{abstract}
\section{Introduction}

The estimation of Average Treatment Effect (ATE) and Conditional Average Treatment Effect (CATE) plays a pivotal role across various disciplines, significantly impacting decision-making processes and policy formulation. In medicine, these estimations guide treatment selections and personalized healthcare strategies \cite{hernan2024, glass2013causal, wager2018estimation}. Within the realm of public policy, they inform the design and evaluation of interventions, from education reforms to social welfare programs \cite{imbens2015causal, hill2011bayesian}. In economics, ATE and CATE estimations are crucial for understanding the impacts of economic policies, labor market interventions, and consumer behavior \cite{angrist2008mostly, heckman2007econometric}. 

A fundamental challenge in this field lies in the correct specification of propensity score and outcome models, particularly when employing methods such as Inverse Probability of Treatment Weighting (IPTW) and Doubly-Robust Estimator (or Augmented IPW) to control for confounding factors \cite{hernan2024, robins1994, bang2005}. These methods, while powerful, are sensitive to model misspecification, which can lead to biased estimates and potentially misleading conclusions \cite{kang2007demystifying, funk2011doubly}. The complexity of real-world scenarios, characterized by high-dimensional data and complex causal relationships, further exacerbates this challenge, necessitating more sophisticated and robust approaches to causal inference \cite{chernozhukov2018double, wager2018estimation}.

The integration of machine learning (ML) methods into causal inference has opened new avenues for addressing complex causal relationships in high-dimensional settings. Athey and Imbens \cite{athey2016recursive} introduced causal trees and forests, adapting random forest algorithms to estimate heterogeneous treatment effects with valid statistical inference. Expanding on this, Wager and Athey \cite{wager2018estimation} developed generalized random forests, extending forest-based methods to a broader class of causal parameters. These approaches have shown promise in settings with many covariates and potential treatment effect heterogeneity. Concurrently, Chernozhukov et al. \cite{chernozhukov2018double} proposed the double machine learning framework, combining flexible ML methods with orthogonalization techniques to achieve valid inference on treatment effects in high-dimensional settings.
Deep learning methods have also made significant inroads in causal inference, offering powerful tools for modeling complex relationships. Shalit et al. \cite{shalit2017estimating} introduced representation learning techniques for estimating individual treatment effects, using neural networks to learn balanced representations of covariates, addressing the fundamental problem of unobserved counterfactuals. The emergence of graph neural networks (GNNs) has further expanded the possibilities in causal inference, particularly for networked data.  Ma et al. \cite{Ma2020CausalIU} demonstrated how GNN-based approaches can estimate heterogeneous treatment effects in the presence of spillover effects, capturing complex dependencies in networked experiments. Recent work has also explored transformer architectures for causal inference tasks. Melnychuk et al. \cite{pmlr-v162-melnychuk22a} introduced the Causal Transformer for estimating counterfactual outcomes over time, effectively capturing long-range dependencies in longitudinal data. Zhang et al. \cite{zhang2023exploring} proposed TransTEE, a transformer-based model for Heterogeneous Treatment Effect (HTE) estimation that handles various types of treatments. Zhang et al. \cite{zhang2023towards} developed Causal Inference with Attention (CInA), enabling zero-shot causal inference on unseen tasks with new data. These deep learning methods offer new ways to handle the challenges of high-dimensional data and complex causal structures in modern causal inference problems.  

Despite significant advancements, current machine learning (ML) and deep learning (DL) approaches to causal inference face notable challenges. A primary limitation is their ability to simultaneously model complex relationships and incorporate structural causal knowledge. Many existing methods excel at flexible modeling of either the outcome regression or propensity score model, but rarely both concurrently. Moreover, they often lack natural mechanisms to explicitly integrate causal knowledge into the learning process. In addition, a particularly persistent challenge in the field is the incorporation of unmeasured confounding into modern DL models, such as transformers \citep{pmlr-v162-melnychuk22a,zhang2023exploring,zhang2023towards}. 

To address these limitations, we propose a novel approach that harnesses the power of transformer models while explicitly incorporating causal structure through a DAG-aware attention mechanism. Our method enables the estimation of crucial causal quantities including the propensity score model $P(A|\mathbf{X})$, the outcome regression model $P(Y|A, \mathbf{X})$, and the bridge function $h(A, W, X)$. Here, $A$ represents the treatment, $\mathbf{X}$ denotes observed confounders, $Y$ is the outcome, and $W$ serves as a proxy for the outcome in scenarios with unmeasured confounding.
This approach allows for seamless integration of these estimated models into IPTW, doubly robust estimators, and proximal inference methods. By doing so, our work bridges the gap between cutting-edge machine learning techniques and classical causal inference methods, offering a more comprehensive framework for causal analysis in complex, real-world scenarios.

The key contributions of this paper are:

\begin{itemize}
    \item Development of a DAG-aware Transformer model that explicitly incorporates causal structure into the attention mechanism, allowing for more accurate modeling of causal relationships in various estimation frameworks including G-formula, IPW, and AIPW.
    
    \item Demonstration of the model's versatility in both joint and separate training paradigms for propensity score and outcome models, revealing context-dependent benefits: joint training excels in ATE estimation, while separate training shows advantages in CATE estimation.
    
    \item Empirical evaluation of the proposed method on diverse datasets (Lalonde-CPS, Lalonde-PSID, ACIC, and Demand), showcasing its superior performance in estimating both average and conditional average treatment effects across various scenarios and sample sizes.
    
    \item Extension of the DAG-aware Transformer to proximal causal inference, demonstrating significant improvements over existing methods, particularly in the NMMR-U framework across different training set sizes.
\end{itemize}

\section{Preliminaries}
\subsection{ATE and CATE}
Consider treatment $A$ and its effect on outcome $Y$. Let $\mathbf{X}$ denote a vector of \textit{observed} confounders. We define $Y^a$ as the counterfactual outcome for each individual had they received ($a=1$) or not received ($a=0$) the treatment. The Average Treatment Effect (ATE), denoted as $\tau$, is then defined as $\tau = \mathbb{E}[Y^{1} - Y^{0}]$. 

While the ATE provides an overall measure of the treatment effect across the entire population, in many cases, it's important to understand how the treatment effect varies across different subgroups or individuals. The CATE, denoted as $\tau(x)$, measures the average treatment effect for a subpopulation with a specific set of covariates $X = x$: $\tau(x) = \mathbb{E}[Y^{1} - Y^{0} | X = x]$. The CATE allows us to capture heterogeneity in treatment effects across different subgroups defined by their covariate values. It's particularly useful in personalized medicine, targeted policy interventions, and other scenarios where the effect of a treatment may vary substantially across different segments of the population.

\subsection{Confounding Control Methods assuming Unconfoundedness}
In causal inference, several methods have been developed to control for \textit{observed} confounding and estimate treatment effects. Our paper focuses primarily on three methods: Standardization (G-formula), Inverse Probability of Treatment Weighting (IPTW) and Augmented Inverse Probability Weighting (AIPW), a form of Doubly Robust estimator \citep{hernan2024}. 

\begin{enumerate}

    \item \textbf{Standardization (G-formula)}:
    Standardization, also known as the G-formula, estimates the ATE by modeling the outcome as a function of treatment and confounders. It then averages over the confounder distribution to estimate the population-level effect. The ATE is estimated as:
    \begin{equation}
        \tau_{G} = \mathbb{E}_X[\mathbb{E}[Y|A=1,X] - \mathbb{E}[Y|A=0,X]]
    \label{eq:gformula}
    \end{equation}
    where $\mu(a,X) = \mathbb{E}[Y|A=a,X]$ is the conditional expectation of the outcome given treatment $a$ and confounders $X$. This method is effective when the outcome model is correctly specified.

    \item \textbf{Inverse Probability of Treatment Weighting (IPTW)}:
    IPTW uses the propensity score to create a pseudo-population in which the treatment assignment is independent of the measured confounders. The ATE is estimated as:
    \begin{equation}
         \tau_{IPTW} = \mathbb{E}\Big[\frac{A Y}{\pi(X)} - \frac{(1-A) Y}{1-\pi(X)}\Big]
    \label{eq:iptw}
    \end{equation}
    where $\pi(X) = P(A=1|X)$ is the propensity score. This method is effective when the propensity score model is correctly specified.
    
    \item \textbf{Augmented Inverse Probability Weighting (AIPW)}:
    AIPW combines IPTW with an outcome regression model, providing robustness against misspecification of either the propensity score model or the outcome model. The ATE is estimated as:
    \begin{equation}
        \begin{aligned}
            \tau_{AIPW} = \mathbb{E}\Big[&\big(\mu(1,X) + \frac{A}{\pi(X)}(Y-\mu(1,X))\big) \\
            &- \big(\mu(0,X) + \frac{1-A}{1-\pi(X)}(Y-\mu(0,X))\big) \Big]
        \end{aligned}
        \label{eq:aipw}
    \end{equation}
    where $\mu(a,X) = \mathbb{E}[Y|A=a, X]$ is the outcome regression function.
\end{enumerate}

\subsection{Proximal Inference}

In proximal inference \citep{Tchetgen2024}, we aim to estimate the expected potential outcome $\mathbb{E}[Y^{a}]$ for each treatment level $a$, in the presence of unobserved confounders $U$, given a set of proxies $(W, Z)$ and observed confounders $X$. The key assumptions are:

\begin{assum} \label{a:indep}
Given $(A, U, W, X, Y, Z)$, $Y \indep Z | A, U, X$ and $W \indep (A, Z) | U, X$.
\end{assum}

\begin{assum} \label{a:comp U}
For all $f \in L^2$ and all $a \in \mathcal{A}, x \in \mathcal{X}$, $\mathbb{E}[ f(U) | A=a, X=x, Z=z ] = 0$ for all $z \in \mathcal{Z}$ if and only if $f(U) = 0$ almost surely.
\end{assum}

\begin{assum} \label{a:comp Z}
For all $f \in L^2$ and all $a \in \mathcal{A}, x \in \mathcal{X}$, $\mathbb{E}[ f(Z) | A=a, W=w, X=x ] = 0$ for all $w \in \mathcal{W}$ if and only if $f(Z) = 0$ almost surely.
\end{assum}

Under these assumptions, there exists a bridge function $h$ satisfying:

\begin{equation}\label{eq:integral_equation}
    \mathbb{E}[Y|A=a, X=x, Z=z] = \int_\mathcal{W} h(a,w,x)p(w|a,x,z)dw
\end{equation}

The expected potential outcomes are given by:

\begin{equation}\label{eq:potential_outcome}
    \mathbb{E}[Y^{a}] = \mathbb{E}_{W,X}[h(a, W, X)]
\end{equation}

The ATE then can be derived from the empirical mean of $\hat{h}$ with $a$ fixed to the value of interest, $\hat{\mathbb{E}}[Y^{a}]=\frac{1}{M}\sum_{i=1}^M \hat{h}(a, w_i, x_i)$.

\section{Methodology}
We propose a novel DAG-aware Transformer model for causal effect estimation that explicitly incorporates causal structure into the attention mechanism. Our approach is flexible and can accommodate various causal scenarios, including those with or without unmeasured confounding.

Given a dataset of $N$ observations, we define a set of possible input nodes. These include: $A$, the treatment variable; $\mathbf{X}$, the observed confounding variables; $U$, representing unmeasured confounding variables; $Y$, the outcome variable; $Z$, the proxy variable for treatment; and $W$, the proxy variable for outcome. 
The specific combination of input nodes used depends on the causal structure being modeled. The output nodes of our model vary based on the estimation method employed:

\begin{itemize}
    \item For standardization or proximal inference: $\hat{Y}$ (estimated outcome), which is $\hat{\mu}(a,X)$ for standardization or $\hat{h}(a, W, X)$ for proximal inference.
    \item For Inverse Probability of Treatment Weighting (IPTW): $\hat{A} = \hat{\pi}(X)$ (estimated propensity score)
    \item For Augmented Inverse Probability Weighting (AIPW): Both $\hat{A}$ and $\hat{Y}$
\end{itemize}

This flexible framework allows our DAG-aware Transformer to adapt to different causal inference scenarios and estimation techniques while maintaining its core structure.

After estimating $\hat{A}$ and $\hat{Y}$, we can plug these values into the corresponding equations to estimate the Average Treatment Effect (ATE) or Conditional Average Treatment Effect (CATE). For instance:

\begin{itemize}
    \item For standardization: ATE = $\mathbb{E}_X[\hat{\mu}(1,X) - \hat{\mu}(0,X)]$
    \item For IPTW: ATE = $\mathbb{E}[\frac{AY}{\hat{\pi}(X)} - \frac{(1-A)Y}{1-\hat{\pi}(X)}]$
    \item For AIPW: ATE = $\mathbb{E}[(\hat{\mu}(1,X) - \hat{\mu}(0,X)) + \frac{A(Y-\hat{\mu}(1,X))}{\hat{\pi}(X)} - \frac{(1-A)(Y-\hat{\mu}(0,X))}{1-\hat{\pi}(X)}]$
    \item For proximal inference: ATE = $\mathbb{E}[\hat{h}(1, W, X) - \hat{h}(0, W, X)]$
\end{itemize}

For CATE estimation, we can condition on specific values of $X$ in these equations. This approach allows us to estimate both population-level and subgroup-level causal effects using our DAG-aware Transformer model.

\subsection{DAG-aware Transformer Architecture}
\begin{wrapfigure}{l}{0.5\textwidth}
\centering
\begin{tikzpicture}[
    node distance=0.8cm and 1.3cm,
    box/.style={rectangle, draw, minimum width=2cm, minimum height=1cm},
    bigbox/.style={rectangle, draw, minimum width=5.5cm, minimum height=6cm},
    attentionbox/.style={rectangle, draw, dashed, fill=yellow!20, minimum width=5cm, minimum height=5.5cm},
    arrow/.style={->, >=stealth, thick},
    state/.style={rectangle, draw, minimum size=0.8cm}
]

\node[box] (input) {Input Nodes ($A$, $\mathbf{X}$, $Y$)};

\node[box, below= 0.5cm of input] (embedding) {Embedding Layer};

\node[bigbox, below=0.5cm of embedding] (transformer) {};

\node[above=-0.5cm of transformer.north](transformer-text) {Transformer Encoder};

\node[attentionbox, below=0.5cm of transformer.north] (attention) {};

\node[above=1cm of attention.center] (dag) {
    \begin{tikzpicture}[node distance=0.8cm, transform shape, scale=0.8]
        \node[state] (X) at (0,0) {$\mathbf{X}$};
        \node[state] (A) at (2,0) {$A$};
        \node[state] (Y) at (4,0) {$Y$};
        \draw[arrow] (X) -- (A);
        \draw[arrow] (X) to[bend right=20] (Y);
        \draw[arrow] (A) -- (Y);
    \end{tikzpicture}
};
\node[above=0.1cm of dag] {Causal DAG};

\node[below=1cm of dag] (adjmatrix-tab) {
    \begin{tabular}{c|ccc}
        & $X$ & $A$ & $Y$ \\ \hline
        $X$ & 0 & 1 & 1 \\
        $A$ & 0 & 0 & 1 \\
        $Y$ & 0 & 0 & 0 \\
    \end{tabular}
};
\node[above=0.1cm of adjmatrix-tab] (adjmatrix) {Adjacency Matrix};

\node[above=0.1cm of attention.south](dag-attention) {DAG-aware Attention};

\node[box, below= 0.5cm of transformer] (mlp) {MLP};

\node[box, below= 0.5cm of mlp] (output) {Output Nodes ($\hat{A}$, $\hat{Y}$)};

\draw[arrow] (input) -- (embedding);
\draw[arrow] (embedding) -- (transformer);
\draw[arrow] (transformer) -- (mlp);
\draw[arrow] (input) -- ++(4,0) |- (mlp);
\draw[arrow] (mlp) -- (output);

\node[right=0.1cm of mlp, text width=2cm, align=left, font=\footnotesize] {Concatenation of encoder output and raw input};


\end{tikzpicture}
\caption{ Architecture of the DAG-aware Transformer model. The Transformer Encoder incorporates the DAG-aware attention mechanism (highlighted with dashed lines), which utilizes the causal structure represented by the DAG. The adjacency matrix derived from the causal DAG informs the DAG-aware attention computation. The model combines the output from the transformer encoder with the raw input through a weighted average, which is then processed by an MLP to produce the final output. For simplicity, layer normalization and feed-forward networks within the transformer encoder are not shown.}
\label{fig:model-architecture}
\end{wrapfigure}
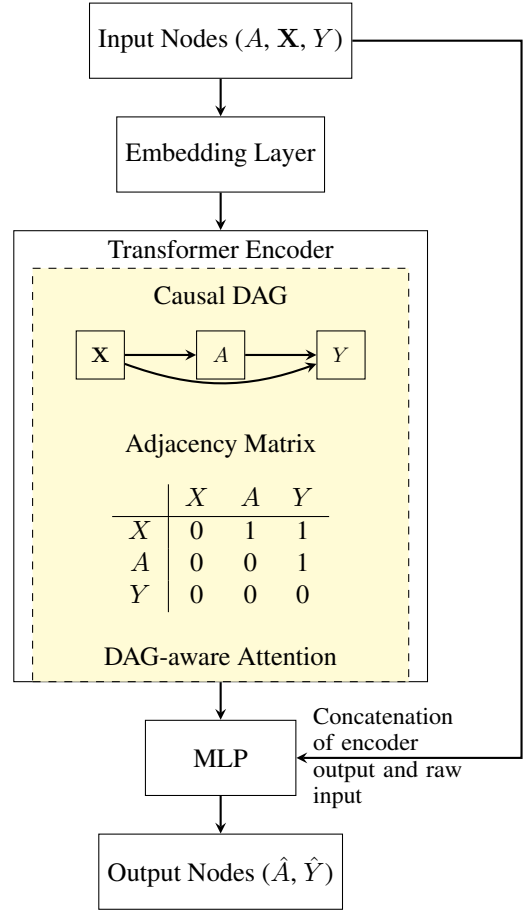

See Figure \ref{fig:model-architecture} for an illustration of our model architecture. We encode the causal DAG into an adjacency matrix $\mathbf{M}^{adj} \in \{0,1\}^{D \times D}$, where $D$ is the number of nodes. Each element $M^{adj}_{ij} = 1$ indicates that there is a directed edge from node $i$ to node $j$. 
We then transform this into an attention mask $\mathbf{M}$:
\begin{equation}
M_{ij} = 
\begin{cases} 
    0 & \text{if } M^{adj}_{ji} = 1 \text{ or } i = j \\
    1 & \text{otherwise}
\end{cases}
\end{equation}
This mask ensures attention flows only along causal pathways and allows self-attention for each node.

Our key innovation lies in incorporating the causal structure into the attention mechanism. In each multi-head attention layer, we compute attention scores $\mathbf{A} = \frac{\mathbf{Q}\mathbf{K}^T}{\sqrt{E}}$ and apply the DAG-based mask $\mathbf{A}^{mask} = \mathbf{A} + \mathbf{M} \cdot (-\infty)$, where $\mathbf{Q}, \mathbf{K} \in \mathbb{R}^{N \times D \times E}$ are the query and key matrices, and $E$ is the embedding dimension. This operation effectively sets attention scores to zero (after softmax) for node pairs not causally linked in the DAG.

The masked attention scores are then normalized using softmax and used to compute the output:
$
\text{Attention}(\mathbf{Q}, \mathbf{K}, \mathbf{V}) = \text{softmax}(\mathbf{A}^{mask})\mathbf{V}
$
where $\mathbf{V}$ is the value matrix.

After the transformer encoder processes the input, we introduce a novel component to our architecture. We concatenate the output from the encoder (weighted by $alpha$) with the raw input before passing it to a Multi-Layer Perceptron (MLP). This approach is inspired by residual connections used in ResNet architectures \cite{he2016deep}. The weighted output is computed as follows: $
\mathbf{Z} = \alpha \cdot \mathbf{H} + \mathbf{X}$ where $\mathbf{H}$ is the output from the Transformer Encoder, $\mathbf{X}$ is the raw input, and $\alpha$ is a hyper parameter. 

The rationale behind this concatenation is to mitigate potential loss of confounding information that might occur when using the encoder alone. We found that relying solely on the encoder output could sometimes result in identical predictions of $\hat{Y}$ regardless of the treatment value, indicating a loss of important confounding information. By incorporating the raw input directly, we ensure that all relevant information is preserved and utilized in the final prediction stage.

The weighted output $\mathbf{Z}$ is then fed into an MLP, which produces the final output:
$
\hat{A}/\hat{Y} = \text{MLP}(\mathbf{Z})
$

This architecture allows our model to leverage both the complex representations learned by the transformer encoder and the raw input features, leading to more robust and accurate causal effect estimations.

\clearpage
\subsection{Model Training and Objective Function}

Our model employs different loss functions depending on the causal inference method used. We present the loss functions for standardization (G-formula), Inverse Probability of Treatment Weighting (IPTW), Augmented Inverse Probability Weighting (AIPW), and proximal inference. Here we assume $Y$ (outcome) is continuous, and $A$ (treatment) is a binary variable. 

\subsubsection{Standardization}
For standardization, we use Mean Squared Error (MSE) loss for maximum likelihood estimation:

\begin{equation}
    \mathcal{L}_G = \text{MSE}(\hat{Y}, Y) = \frac{1}{n} \sum_{i=1}^n (\hat{Y}_i - Y_i)^2
\end{equation}

where $\hat{Y}$ are the model outputs and $Y$ are the true labels.

\subsubsection{Inverse Probability of Treatment Weighting (IPTW)}
For IPTW, we use Binary Cross Entropy (BCE) loss for treatment/propensity score estimation:

\begin{equation}
\begin{aligned}
    \mathcal{L}_{\text{IPTW}} = \text{BCE}(\hat{A}, A) = \\
    -\frac{1}{n} \sum_{i=1}^n [A_i \log(\hat{A}_i) + (1-A_i) \log(1-\hat{A}_i)]
\end{aligned}
\end{equation}

where $\hat{A}$ are the model outputs (estimated propensity scores) and $A$ are the true treatment assignments.

\subsubsection{Augmented Inverse Probability Weighting (AIPW)}
For AIPW, we combine MSE loss for outcome prediction and BCE loss for treatment assignment:

\begin{equation}
    \mathcal{L}_{\text{AIPW}} = \frac{1}{2}(\text{MSE}(\hat{Y}, Y) + \text{BCE}(\hat{A}, A))
\end{equation}

\subsubsection{Proximal Inference}
For proximal inference, following the work of \cite{kompa2022deep}, we introduce two variants: NMMR-U and NMMR-V, based on U-statistics and V-statistics respectively. The empirical risk $\hat{R}_{k,n}$ given data $\mathcal{D} = \{(a_i, w_i, x_i, y_i, z_i)\}_{i=1}^N$ can be written as:

\begin{equation}
    \hat{R}_{k,U,n}(h) = \frac{1}{n(n-1)} \sum_{i,j=1,i\neq j}^n (y_i - h_i)(y_j - h_j)k_{ij}
\end{equation}

\begin{equation}
    \hat{R}_{k,V,n}(h) = \frac{1}{n^2} \sum_{i,j=1}^n (y_i - h_i)(y_j - h_j)k_{ij}
\end{equation}

where $h_i = h(a_i, w_i, x_i)$ and $k_{ij} = k((a_i, z_i, x_i), (a_j, z_j, x_j))$.

To prevent overfitting, we add an L2 penalty $\Lambda[h, \theta_h] = \sum_i \theta_{h,i}^2$, where $\theta_h$ are the model parameters. The penalized risk functions are:

\begin{equation}
    \hat{R}_{k,U,\lambda,n}(h) = \hat{R}_{k,U,n}(h) + \lambda\Lambda[h, \theta_h]
\end{equation}

\begin{equation}
    \hat{R}_{k,V,\lambda,n}(h) = \hat{R}_{k,V,n}(h) + \lambda\Lambda[h, \theta_h]
\end{equation}

The final loss function for training the neural networks is:

\begin{equation}
    \mathcal{L}_{proximal} = (Y - h(A,W,X))^T K (Y - h(A,W,X)) + \lambda\Lambda[h, \theta_h]
\end{equation}

where $(Y - h(A,W,X))$ is a vector of residuals from the neural network's predictions and $K$ is a kernel matrix with entries $k_{ij}$. We use an RBF kernel for $k$. For the U-statistic variant (NMMR-U), we set the main diagonal of $K$ to zero, while for the V-statistic variant (NMMR-V), we include the main diagonal elements.

\subsection{Hyperparameter Tuning and Model Selection}

Hyperparameter tuning and model selection present unique challenges in causal inference, as unlike traditional machine learning tasks with observed labels and cross-validation procedures, we cannot directly observe counterfactual potential outcomes \citep{saito2020counterfactual, mahajan2024empirical}. To address this issue, several methods have been developed \citep{Nie2017QuasioracleEO, saito2020counterfactual}. In our work, we adopt the approach proposed by \citep{saito2020counterfactual}, which aligns well with our training scheme and offers a straightforward implementation.

Given the unobservable nature of true causal effects, we estimate surrogate metrics that approximate these effects. Our procedure is as follows:

\begin{enumerate}
    \item We train a plug-in estimator $\hat{\tau}$ on the validation set using generalized random forests \citep{wager2018estimation}.
    \item We then select models and tune hyperparameters by finding estimators $\tilde{\tau}$ that minimize the difference between $\hat{\tau}$ and $\tilde{\tau}$.
\end{enumerate}

Formally, we select the optimal estimator $\tilde{\tau}^*$ according to
$\tilde{\tau}^{\ast} = \argmin_{\tilde{\tau} \in \mathcal{T}} \, \text{NRMSE}\left(\hat{\tau}, \tilde{\tau}\right)$, where $\mathcal{T}$ is the set of candidate estimators, and Normalized Root Mean Squared Error (NRMSE) is defined as
$\text{NRMSE} = \sqrt{\frac{\frac{1}{n-1}\sum_{i=1}^n (\hat{\tau}(X_i) - \tilde{\tau}(X_i))^2}{\hat{V}(\hat{\tau}(X))}}$.
 Here, $\{\tilde{\tau}(X_i)\}_{i=1}^n$ is a set of ATE or CATE predictions by $\tilde{\tau}(\cdot)$, and $\hat{V}(\hat{\tau}(X))$ is the empirical variance of the ground-truth ATE or CATE approximated by $\hat{\tau}(\cdot)$. A detailed description of the hyper-parameter tuning process, including all considered value ranges, is available in Appendix \ref{app:hyperparameters}.

\section{Experiments}

We evaluate our proposed DAG-aware Transformer method on four diverse datasets, assessing its performance across different causal inference tasks and estimation methods. We use the LaLonde datasets for Average Treatment Effect (ATE) estimation. The ACIC dataset is employed for Conditional Average Treatment Effect (CATE) estimation, allowing us to assess our model's ability to capture heterogeneous treatment effects. Additionally, we extend our evaluation to proximal inference scenarios using the Demand dataset described in \cite{kompa2022deep}, testing our model's adaptability to more complex causal structures. The causal assumptions for each of these experiments are detailed in Appendix~\ref{appendix:causal-assumptions}.

We implement and compare three main causal estimators using our DAG-aware Transformer architecture: Standardization, which estimates the outcome model directly; Inverse Probability Weighting (IPW), which focuses on estimating propensity scores; and Augmented Inverse Probability Weighting (AIPW), a doubly robust method that combines both outcome modeling and propensity score estimation. For each dataset, we compare the performance of these estimators to assess which approach works best in different scenarios. This comparison allows us to evaluate the flexibility and effectiveness of our DAG-aware Transformer across various causal inference methodologies.

For the AIPW estimator, we conduct an additional experiment to compare two training strategies: separate training, where we train the outcome model and propensity score model independently using our DAG-aware Transformer, and joint training, where we train both models simultaneously within a single DAG-aware Transformer architecture. This comparison aims to determine whether the joint learning of outcome and propensity models offers advantages over separate training in terms of estimation accuracy or computational efficiency. We hypothesize that the joint training approach may benefit from shared representations and potentially capture more nuanced interactions between confounders, treatment, and outcomes.

\subsection{Evaluation metrics}
We use normalized root mean square error (NRMSE) to compare our estimated ATE with the true ATE for LaLonde CPS, LaLonde PSID datasets, and the ACIC dataset (described below). This metric provides a standardized measure of the estimation accuracy, allowing for comparisons across different datasets and methods. In the case of the Demand dataset and proximal inference, we follow the evaluation protocol outlined in \cite{kompa2022deep}. Specifically, performance is evaluated using causal mean squared error (c-MSE) across 10 equally-spaced price points between 10 and 30, comparing estimated potential outcomes $\hat{E}[Y^a]$ against Monte Carlo simulations of the true $E[Y^a]$. This approach allows us to assess the model's ability to estimate counterfactual outcomes across a range of treatment values. 

\subsection{Baseline models}
To evaluate the performance of our DAG-aware Transformer, we compare it against several established baseline models, each chosen for their relevance to causal inference tasks. Our primary baselines include Generalized Random Forests (GRF) \citep{wager2018estimation}, a versatile non-parametric method that adapts random forests for heterogeneous treatment effect estimation, and Multilayer Perceptron (MLP), a standard neural network architecture that serves as a comparison to our more complex transformer-based model. Additionally, for the Inverse Probability Weighting (IPW) estimator, we include a Naive IPW baseline that uses uniform weights, demonstrating the importance of proper weighting in causal inference.

For the Lalonde and ACIC datasets, we implement GRF and MLP for each estimator (G-formula, IPW, and AIPW), with the naive IPW additionally used for the IPW estimator. In the case of the Demand dataset, we employ MLP as our primary baseline due to its effectiveness in handling the dataset's specific characteristics \cite{kompa2022deep}. To ensure a fair comparison, all baseline models (except the naive IPW) underwent careful hyperparameter tuning.

\subsection{Lalonde dataset for ATE estimation}

\begin{figure}[ht]
\vskip 0.2in
\begin{center}
\centerline{\includegraphics[width=\columnwidth]{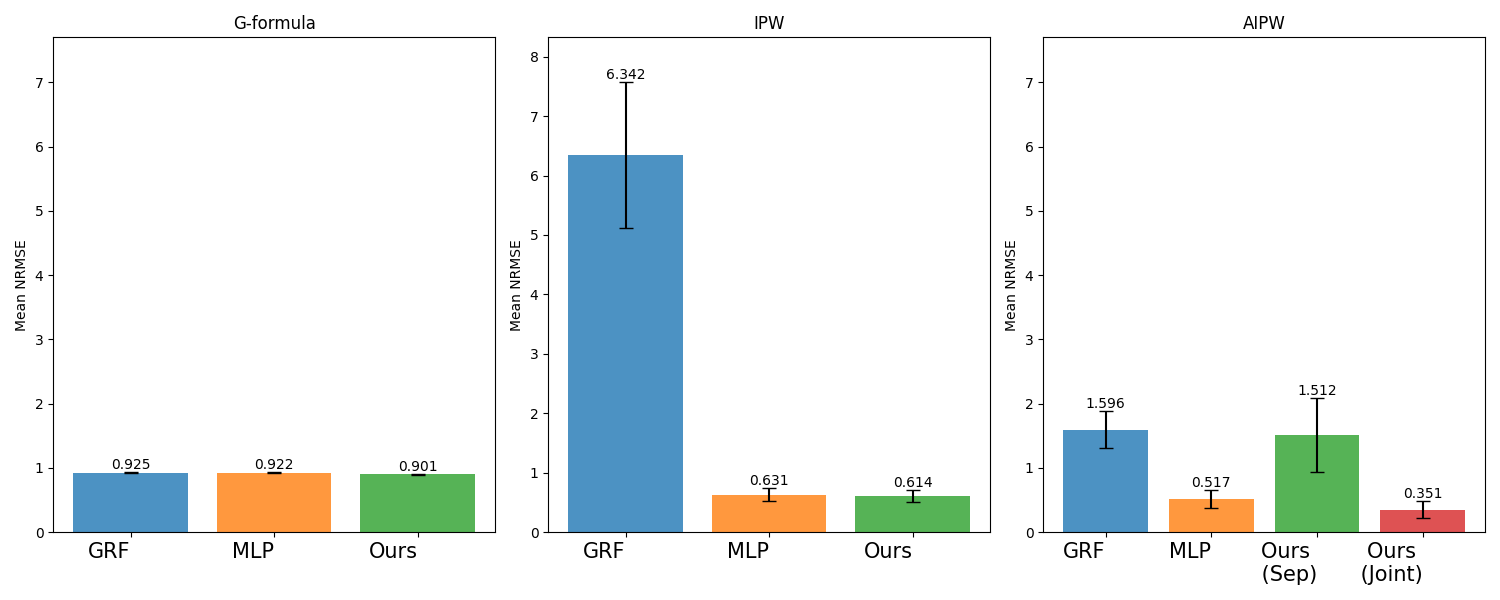}}
\caption{Mean NRMSE with Standard Error for LaLonde CPS Dataset}
\label{fig:lalonde_cps}
\end{center}
\vskip -0.2in
\end{figure}

\begin{figure}[ht]
\vskip 0.2in
\begin{center}
\centerline{\includegraphics[width=\columnwidth]{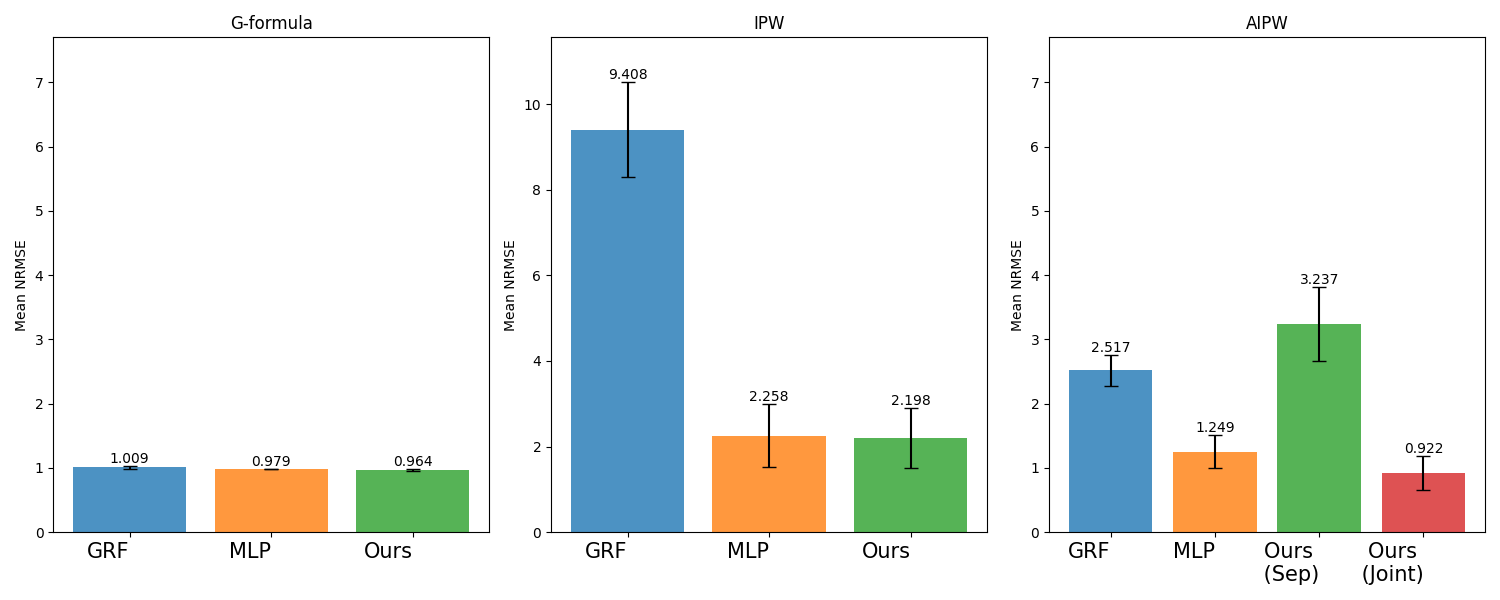}}
\caption{Mean NRMSE with Standard Error for LaLonde PSID Dataset}
\label{fig:lalonde_psid}
\end{center}
\vskip -0.2in
\end{figure}

The LaLonde datasets, derived from the National Supported Work (NSW) Demonstration program, include the Current Population Survey (CPS) and Panel Study of Income Dynamics (PSID) \citep{lalonde1986evaluating}. We define the treatment (A) as participation in the NSW program, the outcome (Y) as earnings in 1978, and the covariates (X) as age, education, race, marital status, earnings in 1974, and earnings in 1975. The LaLonde CPS dataset comprises 15,992 control units and 185 treated units, while the LaLonde PSID dataset consists of 2,490 control units and the same 185 treated units. The true ATE for both datasets is 1,794.34.
To ensure robust evaluation, we create 10 distinct samples from each dataset using bootstrap with replacement with different random seeds. Figures \ref{fig:lalonde_cps} and \ref{fig:lalonde_psid} present the Normalized Root Mean Squared Error (NRMSE) results with standard errors for the LaLonde CPS and PSID datasets, respectively. Table \ref{tab:lalonde_results} in Appendix \ref{appendix:lalonde_results} provide the corresponding numerical values.

Our DAG-aware Transformer model consistently outperformed the baseline methods across both datasets and all estimation techniques. For the Lalonde-CPS dataset, in G-formula estimation, our Transformer achieved the lowest NRMSE of 0.901, compared to 0.925 for GRF and 0.922 for MLP. The Transformer's performance in IPW was equally impressive, with an NRMSE of 0.614, significantly outperforming GRF (6.342) and slightly better than MLP (0.631). In AIPW, our jointly trained Transformer-Joint model achieved the best performance with an NRMSE of 0.351, substantially outperforming all other methods including the separately trained Transformer-Sep (1.512).

Similar trends were observed in the Lalonde-PSID dataset. For G-formula, our Transformer model achieved the lowest NRMSE of 0.964, compared to 1.009 for GRF and 0.979 for MLP. In IPW estimation, the Transformer (NRMSE 2.198) outperformed both GRF (9.408) and MLP (2.258). For AIPW, the Transformer-Joint model again showed superior performance with an NRMSE of 0.922, significantly better than GRF (2.517), MLP (1.249), and Transformer-Sep (3.237).

Notably, the jointly trained Transformer model (Transformer-Joint) in AIPW consistently outperformed its separately trained counterpart (Transformer-Sep) across both datasets, highlighting the benefits of joint training in capturing complex interactions between confounders, treatment, and outcomes. The superior performance of joint training in ATE estimation can be attributed to several factors. Firstly, joint training allows for shared representation learning, potentially capturing complex interactions between confounders, treatment, and outcomes that might be missed when models are trained separately \cite{shalit2017estimating}. This shared learning can be particularly beneficial when the same features influence both treatment assignment and outcomes. Secondly, joint training can lead to more efficient use of the data, especially in smaller datasets, by leveraging information across tasks \cite{kunzel2019metalearners}. Furthermore, the ATE, being a population-level estimate, may benefit from the regularizing effect of joint training, which can prevent overfitting to noise in individual components \cite{shi2019adapting}. Joint training also allows for end-to-end optimization of the entire causal inference pipeline, potentially leading to a better global optimum \cite{yao2018representation}. Additionally, in scenarios where there's model misspecification, joint training might offer robustness by allowing complementary strengths of the outcome and propensity models to compensate for each other's weaknesses \cite{chernozhukov2018double}. These advantages seem to outweigh the potential benefits of specialized modeling in the ATE context, where the goal is to estimate an average effect rather than heterogeneous effects across subpopulations.

\subsection{ACIC dataset for CATE estimation}
\begin{figure}[ht]
\vskip 0.2in
\begin{center}
\centerline{\includegraphics[width=\columnwidth]{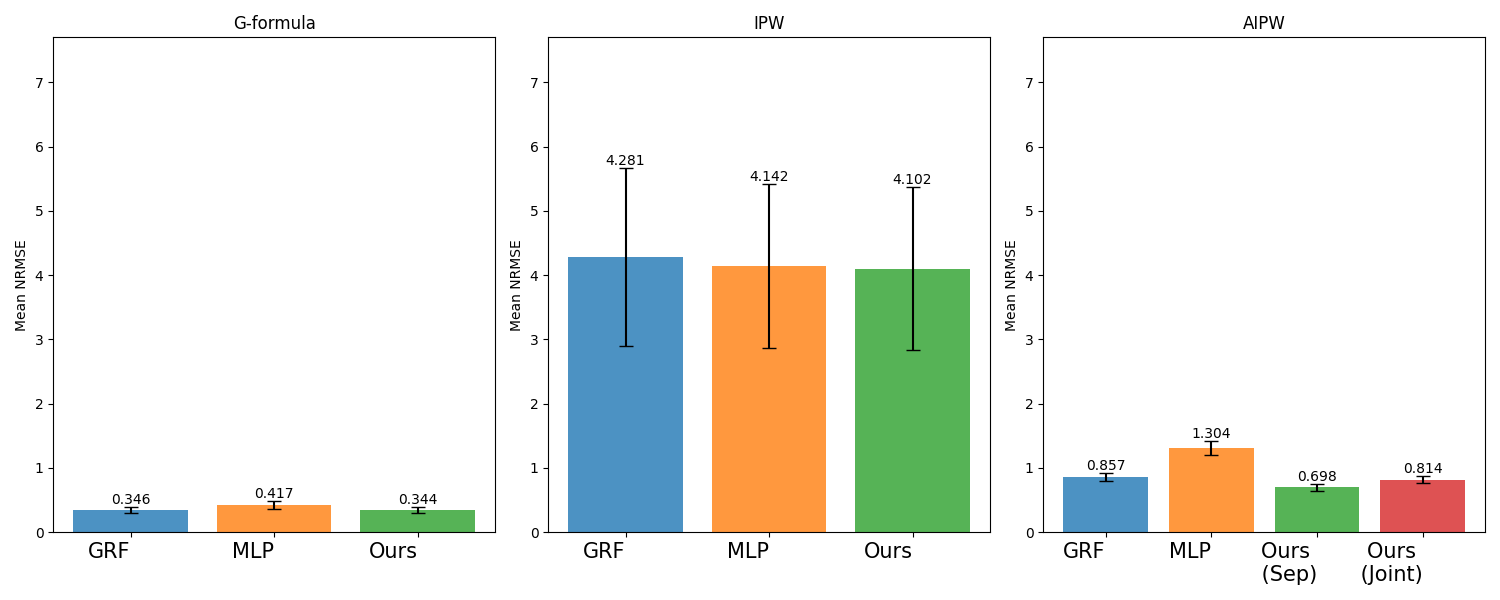}}
\caption{Mean NRMSE with Standard Error for LaLonde PSID Dataset}
\label{fig:acic}
\end{center}
\vskip -0.2in
\end{figure}

The ACIC dataset from the 2016 Atlantic Causal Inference Conference data challenge \citep{dorie2019automated} is based on real covariates with synthetically simulated treatment assignment and potential outcomes. We analyze 10 instances from different data-generating processes, each containing 58 pre-treatment variables, a binary treatment assignment, observed outcome, and ground truth potential outcomes.
Figure \ref{fig:acic} presents the NRMSE with standard errors with respect to the 10 instances of datasets for each method. Table \ref{tab:acic_results} in Appendix \ref{appendix:acic_results} provide the corresponding numerical values.

Our DAG-aware Transformer model demonstrated competitive performance in CATE estimation across all estimation techniques on the ACIC dataset. In the G-formula estimation, our Transformer achieved the lowest NRMSE of 0.344, slightly outperforming GRF (0.346) and showing a substantial improvement over MLP (0.417). This suggests that our model effectively captures the underlying causal structure in estimating conditional treatment effects.

For the IPW estimator, all methods showed higher NRMSE values, indicating the challenging nature of this approach for CATE estimation. Nevertheless, our Transformer model achieved the best performance with an NRMSE of 4.102, marginally better than MLP (4.142) and GRF (4.281). This consistent, albeit small, improvement across different estimation techniques highlights the robustness of our approach.

In the AIPW setting, both variants of our Transformer model outperformed the baseline methods, with the separately trained Transformer (Transformer-Sep) achieving the lowest NRMSE of 0.698, slightly outperforming the jointly trained version (Transformer-Joint) with an NRMSE of 0.814. This contrasts with our ATE estimation results, where joint training was superior. This phenomenon aligns with recent literature on doubly robust estimators for CATE. \cite{Nie2017QuasioracleEO} argue that jointly fitting the outcome and propensity models can lead to overfitting in CATE estimation, particularly when the propensity model is misspecified. They propose separate estimation of nuisance functions to maintain doubly robust properties. Similarly, \cite{kennedy2020optimal} suggests that separate estimation can help avoid bias amplification in CATE settings. The complexity of capturing heterogeneous treatment effects in CATE may benefit from the specialization afforded by separate training, allowing each model to focus on its specific task without interference. These findings underscore the importance of considering both joint and separate training approaches in causal inference tasks, as the optimal strategy may depend on the specific estimation problem and dataset characteristics.

\subsection{Demand for proximal inference}

The goal of this experiment is to estimate the effect of airline ticket price $A$ on sales $Y$, confounded by unobserved demand $U$. We use fuel cost $Z$ as a treatment-inducing proxy and website views $W$ as an outcome-inducing proxy (Figure \ref{fig:demand_dag} in Appendix \ref{appendix:proximal_results}). We train our model on simulated datasets with sample sizes ranging from 1,000 to 50,000.  We use a heldout dataset of 1,000 draws from $W$ to compute predictions. We perform 20 replicates for each sample size. Figure \ref{fig:proximal} summarizes the c-MSE distribution across sample sizes. Table ~\ref{tab:demand_results} in Appendix \ref{appendix:proximal_results} provides the corresponding numerical values. 

\begin{figure}[ht]
\vskip 0.2in
\begin{center}
\centerline{\includegraphics[width=\columnwidth]{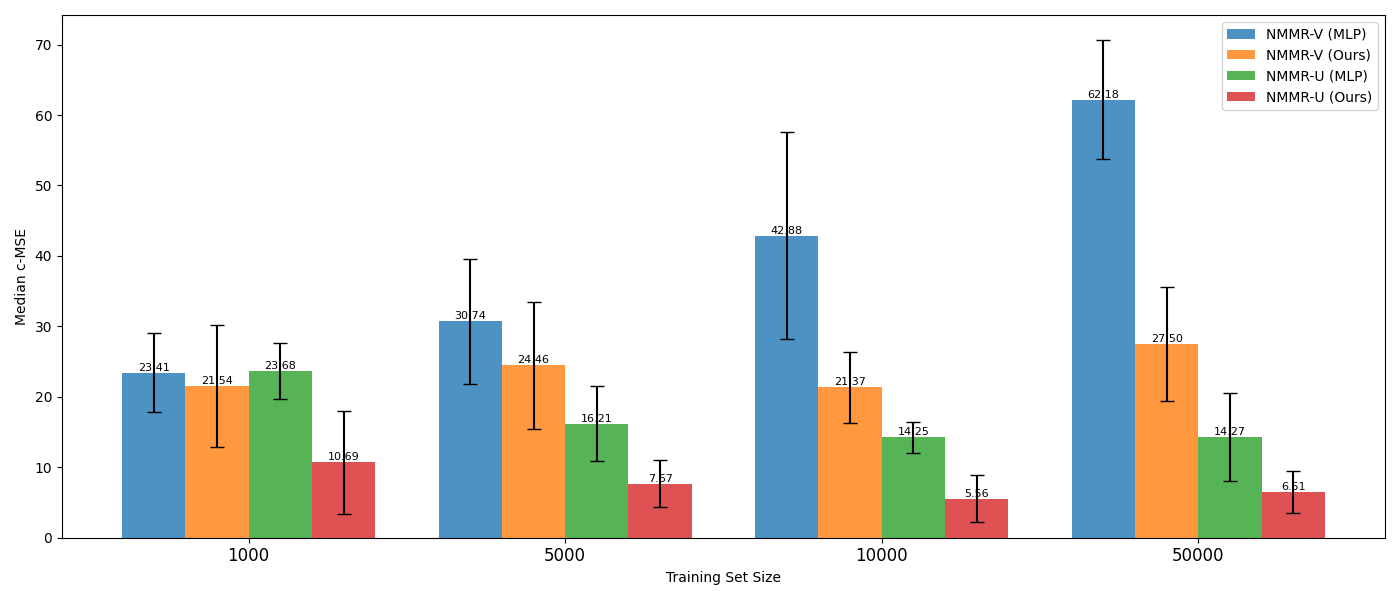}}
\caption{Median (IQR) of c-MSE for Demand Dataset}
\label{fig:proximal}
\end{center}
\vskip -0.2in
\end{figure}

Our Transformer model consistently outperformed the MLP baseline, with the most significant improvements observed when using the NMMR-U estimator. For instance, with a training set size of 10,000, our NMMR-U Transformer achieved a median c-MSE of 5.56 (IQR: 6.72), substantially lower than the MLP's 14.25 (IQR: 4.46). This performance advantage was maintained across all training set sizes, from 1,000 to 50,000 samples. 

\section{Conclusion}
In this paper, we introduced a novel transformer-based approach for causal inference that addresses key limitations in existing methods. Our model's primary innovation lies in its ability to encode any causal DAGs into the attention mechanism, allowing it to handle a wide range of causal scenarios. The causal-aware attention mechanism we developed explicitly models the encoded causal structure, leading to more accurate estimation of treatment effects. Our experimental results demonstrate the effectiveness of our approach across various synthetic and real-world datasets, showing improved performance compared to existing methods. While our work represents a step forward in causal inference using transformer architectures, there are several directions for future research.  First, we recognize the importance of swift adaptability to a wider range of causal structures. Future work can explore developing a more generalized encoding mechanism that can quickly accommodate diverse causal graphs without requiring extensive retraining. This could involve creating a meta-learning framework that learns to adapt to new causal structures efficiently. Additionally, we acknowledge the need to investigate the robustness of our approach against potential noise or misspecifications in the input DAG. Future studies can systematically introduce perturbations to the causal graph to assess how our model's performance degrades under various levels of DAG uncertainty.

\newpage
\bibliography{reference}
\bibliographystyle{icml2025}

\newpage
\appendix
\onecolumn
\section{Causal Assumptions}
\label{appendix:causal-assumptions}
To ensure valid causal inference, several key assumptions must hold. In this paper, we primarily focus on three fundamental assumptions:

\begin{enumerate}
    \item \textbf{Positivity (or Overlap)}: For every $x \in \text{support}(X)$, and $\forall a \in \{0,1\}$, $P(A=a|X=x)>0$. 
    
    This assumption ensures that there is a non-zero probability of receiving each treatment level for all possible values of the observed covariates. It is crucial for estimating treatment effects across the entire covariate space and prevents extrapolation to regions where we have no information about one of the treatment groups.
    
    \item \textbf{Exchangeability (or Unconfoundedness)}: $Y^{a} \perp\!\!\!\perp A | X, \forall a \in \{0,1\}$. 
    
    This assumption implies that, conditional on the observed confounders $X$, the potential outcomes $Y^{a}$ are independent of the treatment assignment $A$. In other words, after controlling for $X$, there are no unmeasured confounders that affect both the treatment assignment and the outcome. This is also known as the "no unmeasured confounding" assumption.
    
    \item \textbf{Consistency}: If $A=a$, then $Y^{a} = Y$. 
    
    This assumption states that the potential outcome under a particular treatment level is the same as the observed outcome if the individual actually receives that treatment level. It ensures that the observed outcomes can be used to estimate the potential outcomes.
\end{enumerate}

For the Lalonde and ACIC experiments, we assume that all three of these assumptions hold. For the proximal inferernce experiment, we remove the strong assumption of no unmeasured confounding (Assumption 2). 

\section{Experiments}

\subsection{LaLonde Dataset Results}
\label{appendix:lalonde_results}

\begin{table}[h]
\centering
\caption{NRMSE for ATE estimation on LaLonde datasets}
\label{tab:lalonde_results}
\begin{tabular}{llccc}
\hline
Dataset & Estimator & Method & Mean NRMSE & SE NRMSE \\
\hline
\multirow{10}{*}{Lalonde-CPS} 
 & \multirow{3}{*}{G-formula} & GRF & 0.925 & 0.002 \\
 & & MLP & 0.922 & 0.006 \\
 & & Transformer & \textbf{0.901} & 0.005 \\
\cline{2-5}
 & \multirow{3}{*}{IPW} & GRF & 6.342 & 1.227 \\
 & & MLP & 0.631 & 0.104 \\
 & & Transformer & \textbf{0.614} & 0.100 \\
\cline{2-5}
 & \multirow{4}{*}{AIPW} & GRF & 1.596 & 0.294 \\
 & & MLP & 0.517 & 0.142 \\
 & & Transformer-Sep & 1.512 & 0.578 \\
 & & Transformer-Joint & \textbf{0.351} & 0.130 \\
\hline
\multirow{10}{*}{Lalonde-PSID} 
 & \multirow{3}{*}{G-formula} & GRF & 1.009 & 0.021 \\
 & & MLP & 0.979 & 0.004 \\
 & & Transformer & \textbf{0.964} & 0.014 \\
\cline{2-5}
 & \multirow{3}{*}{IPW} & GRF & 9.408 & 1.108 \\
 & & MLP & 2.258 & 0.745 \\
 & & Transformer & \textbf{2.198} & 0.705 \\
\cline{2-5}
 & \multirow{4}{*}{AIPW} & GRF & 2.517 & 0.242 \\
 & & MLP & 1.249 & 0.259 \\
 & & Transformer-Sep & 3.237 & 0.571 \\
 & & Transformer-Joint & \textbf{0.922} & 0.264 \\
\hline
\end{tabular}
\end{table}
\newpage
\subsection{ACIC Dataset Results}
\label{appendix:acic_results}

\begin{table}[h]
\centering
\caption{NRMSE for CATE estimation on ACIC dataset}
\label{tab:acic_results}
\begin{tabular}{llcc}
\hline
Estimator & Method & Mean NRMSE & SE NRMSE \\
\hline
\multirow{3}{*}{G-formula} 
 & GRF & 0.346 & 0.044 \\
 & MLP & 0.417 & 0.062 \\
 & Transformer & \textbf{0.344} & 0.054 \\
\hline
\multirow{3}{*}{IPW} 
 & GRF & 4.281 & 1.388 \\
 & MLP & 4.142 & 1.278 \\
 & Transformer & \textbf{4.102} & 1.272 \\
\hline
\multirow{4}{*}{AIPW} 
 & GRF & 0.857 & 0.059 \\
 & MLP & 1.304 & 0.112 \\
 & Transformer-Sep & \textbf{0.698} & 0.057 \\
 & Transformer-Joint & 0.814 & 0.052 \\
\hline
\end{tabular}
\end{table}

\subsection{Proximal inference}
\label{appendix:proximal_results}
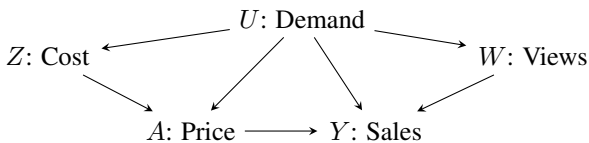
\begin{figure}[ht]
    \centering
    \begin{tikzpicture}[> = stealth, shorten > = 1pt, auto, node distance = 2cm]
    \tikzstyle{every state}=[
        draw = black,
        thick,
        fill = white,
        minimum size = 6mm
    ]
    \node (A) {$A$: Price};
    \node (Z) [above left = 0.5cm and 0.5cm of A] {$Z$: Cost};
    \node (U) [above right = 1cm and -0.2cm of A] {$U$: Demand};
    \node (Y) [right = 1cm of A] {$Y$: Sales};
    \node (W) [above right = 0.5cm and 0.5cm of Y] {$W$: Views};

    \path[->] (A) edge node {} (Y);
    \path[->] (U) edge node {} (A);
    \path[->] (U) edge node {} (Y);
    \path[->] (U) edge node {} (Z);
    \path[->] (U) edge node {} (W);
    \path[->] (Z) edge node {} (A);
    \path[->] (W) edge node {} (Y);
    \end{tikzpicture}
    \caption{Causal DAG for the Demand experiment.}
    \label{fig:demand_dag}
\end{figure}

\begin{table}[h]
\centering
\caption{Median c-MSE (IQR) for Demand dataset across different training set sizes}
\label{tab:demand_results}
\begin{tabular}{llcccc}
\hline
Method & Model & 1000 & 5000 & 10000 & 50000 \\
\hline
\multirow{2}{*}{NMMR-V} 
 & MLP & 23.41 (11.26) & 30.74 (17.73) & 42.88 (29.45) & 62.18 (16.97) \\
 & Transformer & 21.54 (17.42) & 24.46 (17.93) & 21.37 (10.12) & 27.50 (16.30) \\
\hline
\multirow{2}{*}{NMMR-U} 
 & MLP & 23.68 (8.02) & 16.21 (10.55) & 14.25 (4.46) & 14.27 (12.47) \\
 & Transformer & \textbf{10.69 (14.72)} & \textbf{7.67 (6.70)} & \textbf{5.56 (6.72)} & \textbf{6.51 (5.90)} \\
\hline
\end{tabular}
\end{table}

\section{Hyperparameter Tuning}
\label{app:hyperparameters}

We performed extensive hyperparameter tuning for each dataset and estimation method. The hyperparameter spaces explored are detailed below. For each configuration, we report the range of values considered during the tuning process.

\begin{table}[h]
\centering
\caption{Hyperparameter tuning ranges for Lalonde-CPS dataset}
\label{tab:lalonde_cps_hyperparams}
\begin{tabular}{lccc}
\hline
Parameter & G-formula & IPW & AIPW \\
\hline
Number of epochs & 80 & 20 & 20 \\
Batch size & 32 & 32 & 32 \\
Learning rate & 0.001 & 0.001 & 0.001 \\
L2 penalty & 3e-05, 3e-03 & 3e-05, 3e-03 & 3e-05, 3e-03 \\
Network width (MLP) & 80 & 80 & 80 \\
Input layer depth (MLP) & 2--4 & 1--2 & 2--6 \\
Number of layers (encoder) & 2--4 & 1--2 & 1--2 \\
Dropout rate & 0.0001 & 0.0001--0.001 & 0.0001 \\
Embedding dimension (encoder) & 40 & 40 & 40 \\
Feedforward dimension (encoder) & 80 & 80 & 80 \\
Number of heads (encoder) & 2 & 1--2 & 1--2 \\
Encoder weight (alpha) & 0.02 & 0.002--0.02 & 0.02 \\
\hline
\end{tabular}
\end{table}

\begin{table}[h]
\centering
\caption{Hyperparameter tuning ranges for Lalonde-PSID dataset}
\label{tab:lalonde_psid_hyperparams}
\begin{tabular}{lccc}
\hline
Parameter & G-formula & IPW & AIPW \\
\hline
Number of epochs & 100 & 30 & 30 \\
Batch size & 32, 64 & 64 & 32, 64 \\
Learning rate & 0.001--0.01 & 0.001 & 0.001 \\
L2 penalty & 3e-05 & 3e-05 & 3e-05--3e-03 \\
Network width (MLP) & 80 & 10, 20 & 40 \\
Input layer depth (MLP) & 6--16 & 1 & 4--8 \\
Number of layers (encoder) & 1--2 & 1 & 1--2 \\
Dropout rate & 0.0001 & 0.0001 & 0.0001 \\
Embedding dimension (encoder) & 40 & 10--40 & 20--40 \\
Feedforward dimension (encoder) & 80 & 20--80 & 40--80 \\
Number of heads (encoder) & 1--2 & 1 & 2 \\
Encoder weight (alpha) & 0.002--0.02 & 0.02--0.2 & 0.02--0.2 \\
\hline
\end{tabular}
\end{table}

\begin{table}[h]
\centering
\caption{Hyperparameter tuning ranges for ACIC dataset}
\label{tab:acic_hyperparams}
\begin{tabular}{lccc}
\hline
Parameter & G-formula & IPW & AIPW \\
\hline
Number of epochs & 500 & 30 & 500 \\
Batch size & 64 & 64, 128, 256 & 64, 128, 256 \\
Learning rate & 1e-03 & 1e-03 & 1e-04 \\
L2 penalty & 3e-08 & 3e-05 & 3e-08, 3e-04 \\
Network width (MLP) & 40 & 40, 60, 80 & 40, 60, 80, 120 \\
Input layer depth (MLP) & 8--16 & 4--16 & 2--16 \\
Number of layers (encoder) & 16 & 2--4 & 2--8 \\
Dropout rate & 0.0001 & 0.0001 & 0.0001--0.003 \\
Embedding dimension (encoder) & 256 & 40, 60, 80 & 256 \\
Feedforward dimension (encoder) & 1024 & 80, 320 & 512, 1024 \\
Number of heads (encoder) & 4 & 2 & 2--4 \\
Encoder weight (alpha) & 0.02--0.2 & 0.002--0.2 & 0.1--2 \\
\hline
\end{tabular}
\end{table}

\begin{table}[h]
\centering
\caption{Hyperparameter tuning ranges for Demand dataset (Proximal Inference)}
\label{tab:demand_hyperparams}
\begin{tabular}{lc}
\hline
Parameter & Range \\
\hline
Number of epochs & 1000 \\
Batch size & 32, 64 \\
Learning rate & 0.001 \\
L2 penalty & 3e-06 \\
Network width (MLP) & 160 \\
Input layer depth (MLP) & 8, 16 \\
Number of layers (encoder) & 1, 2 \\
Dropout rate & 0 \\
Embedding dimension (encoder) & 40 \\
Feedforward dimension (encoder) & 40, 80 \\
Number of heads (encoder) & 1, 2 \\
Encoder weight (alpha) & 0.001--0.025 \\
\hline
\end{tabular}
\end{table}

For each dataset and estimation method, we performed a grid search over these hyperparameter spaces. The final model for each configuration was selected based on the best performance on a held-out validation set.

\end{document}